\documentclass[11pt,a4paper]{article}
\usepackage[hyperref]{acl2017}
\usepackage{times}
\usepackage{latexsym}
\usepackage{flushend}
\usepackage{url}

\aclfinalcopy 
\def\aclpaperid{138} 

\title{
  UW-FinSent at SemEval-2017 Task 5:\\ Sentiment Analysis on Financial News Headlines\\using Training Dataset Augmentation
}

\author{
  Vineet John \\
  University of Waterloo \\
  {\tt vineet.john@uwaterloo.ca} \\
  \And
  Olga Vechtomova \\
  University of Waterloo \\
  {\tt ovechtomova@uwaterloo.ca} \\
}

\date{}

\begin{document}

\maketitle

\begin{abstract}
  This paper discusses the approach taken by the UWaterloo team to arrive at a solution for the Fine-Grained Sentiment Analysis problem posed by Task 5 of SemEval 2017. The paper describes the document vectorization and sentiment score prediction techniques used, as well as the design and implementation decisions taken while building the system for this task. The system uses text vectorization models, such as N-gram, TF-IDF and paragraph embeddings, coupled with regression model variants to predict the sentiment scores. Amongst the methods examined, unigrams and bigrams coupled with simple linear regression obtained the best baseline accuracy. The paper also explores data augmentation methods to supplement the training dataset. This system was designed for Subtask 2 (News Statements and Headlines).
\end{abstract}

\section{Introduction} 
\label{sec:introduction}
The goal of this SemEval task is to identify fine-grained levels of sentiment polarity in financial news headlines and microblog posts. Specifically, the task aims at identifying bullish (optimistic) sentiment, expressing the belief that the stock price will increase, and bearish (pessimistic) sentiment, expressing the belief that the stock price will decline. The expressed sentiment is quantified as floating point values in the range of -1 (very negative/bearish) to 1 (very positive/bullish), with 0 denoting neutral sentiment. \cite{cortis-EtAl:2017:SemEval}. This paper describes our system developed for subtask 2 (News Statements and Headlines).

While developing the system for this subtask, we systematically evaluated a number of alternative solutions for each step in the pipeline. Specifically, we investigated different document vectorization approaches, such as N-gram models, TF-IDF and paragraph vectors. A number of regression models were evaluated, namely, Simple Linear Regression, Support Vector Regression and XGBoost Linear Regression.

One of the challenges with performing sentiment analysis in the financial domain is scarcity of training data. We explored different approaches to augment the training data provided by the task organizers with training data from other sources in the financial domain, as well as using out-of-domain sentiment resources.

\section{Approach} 
\label{sec:approach}
  The overall approach to predicting the sentiment of the test dataset headlines is detailed below.
  \begin{itemize}
    \item 
      \textbf{Pre-Processing \& Cleaning}

      This step is needed to simplify and sanitize the input set of headlines. In the context of this task, since the headlines were short snippets ranging from 5 to 15 words in length, the only pre-processing done was replacing the name of the organization being spoken of in the headlines, with a generic organization name, to reduce the feature space.
      
    \item 
      \textbf{Text Vectorization}

      The objective is to vectorize the textual content of the headlines into a numeric representation that a statistical learning model can then be trained on. N-gram models, TF-IDF and Paragraph Vector implementations were explored for this purpose. N-gram models generally performed the best on the trial dataset, followed by TF-IDF, and Paragraph Vectors. Of the different N-gram configurations experimented with, word N-grams that used a combination of unigrams and bigrams achieved the best baseline scores. These techniques are further discussed in Section~\ref{sec:document_vectorization}.
      
    \item 
      \textbf{Statistical Model Learning}

      The objective is to use the vector representations of the headlines as features and learn a model to predict the sentiment scores. Simple Linear Regression, Support Vector Regression and XGBoost Linear Regression were the learning methods that were used. The linear regression methods consistently outperformed Support Vector Regression and XGBoost regression in experiments on the training dataset. These techniques are discussed in Section~\ref{sec:regression_models}.
  \end{itemize} 


\begin{table*}[t] 
  \centering
  \begin{tabular}{ | c | c |  c | c | } \hline
    \textbf{Vectorization Method} & \textbf{Learning Model} & \textbf{$R^2$ Score} & \textbf{Cosine Similarity}\\ \hline 
        Unigrams \& Bigrams & Simple Linear Regression & 0.38 & 0.63 \\ 
    \hline
        Unigrams \& Bigrams & Support Vector Regression & 0.38 & 0.63 \\ 
    \hline
        Unigrams \& Bigrams & XGBoost Regression & 0.21 & 0.50 \\ 
    \hline
        TF-IDF & Simple Linear Regression & -0.10 & 0.50 \\ 
    \hline
        TF-IDF & Support Vector Regression & 0.38 & 0.63 \\ 
    \hline
        TF-IDF & XGBoost Regression & 0.19 & 0.47  \\ 
    \hline
        Doc2Vec & Simple Linear Regression & -4.69 & 0.04  \\ 
    \hline
        Doc2Vec & Support Vector Regression & -0.05 & 0.08  \\ 
    \hline
        Doc2Vec & XGBoost Regression & -0.06 & 0.06  \\ 
    \hline
  \end{tabular}
  \caption{Experimental Results}
  \label{tab:semeval_baselines}
\end{table*}

\section{Document Vectorization} 
\label{sec:document_vectorization}
  Document vectorization is needed to convert the text content of the SemEval headlines into a numeric vector representation that can be utilized as features, which can then be used to train a machine learning model on. The methods for vectorization used are listed in the subsections below.

  \subsection{N-gram Model} 
  \label{sub:n_gram_model}
    For the purpose of this task, a vectorizer implementation using Scikit-Learn \cite{pedregosa2011scikit} was used to obtain vector representations of the SemEval headlines, since they have been proven to be an effective representation of textual content for sentiment classification in general \cite{wang2012baselines}.
  

  \subsection{TF-IDF Model} 
  \label{sub:tf_idf_model}
    The TF-IDF implementation in Scikit-Learn \cite{pedregosa2011scikit} was used to obtain vector representations of the SemEval headlines.
  

  \subsection{Paragraph Vector Model} 
  \label{sub:paragraph_vectors_doc2vec}

    A Paragraph Vector representation model is comprised of an unsupervised learning algorithm that learns fixed-size vector representations for variable-length pieces of texts such as sentences and documents \cite{le2014distributed}. The vector representations are learned to predict the surrounding words in contexts sampled from the paragraph. In the context of the SemEval headlines, the vector representations were learned for the complete headline.

    Two distinct implementations were explored while attempting to vectorize the headlines using the Paragraph Vector approach.
    \begin{itemize}
      \item 
        Doc2Vec: A Python library implementation in Gensim\footnote{https://radimrehurek.com/gensim/models/doc2vec.html}.
      \item 
        FastText: A standalone implementation in C++ \cite{bojanowski2016enriching} \cite{joulin2016bag}.
    \end{itemize}

    Doc2Vec was the final choice that was opted for due to the ease of integration into the existing system. The paragraph embeddings for Doc2Vec are trained using the SemEval training headlines corpus.
  


\section{Regression Models} 
\label{sec:regression_models}

  Three different regression implementations were used to train models to predict the sentiment scores of the headlines:
  \begin{itemize}
    \item 
      \textbf{Simple Linear Regression} 

      This is the standard version of linear regression that simply learns the weights for the feature vector that minimize the cost function, which is represented as a Euclidean loss function.

    \item 
      \textbf{Support Vector Regression }

      The idea of SVR is based on the computation of a linear regression function in a high dimensional feature space where the input data is mapped using a non-linear function \cite{basak2007support}.

      Instead of minimizing the observed training error, Support Vector Regression (SVR) attempts to minimize the generalization error bound so as to achieve generalized performance.

    \item 
      \textbf{XGBoost Regression} 

      This is an ensemble method for regression that coalesces several `weak' learners into a single `strong' learner by iteratively minimizing the least squares error or Euclidean loss incurred by the cost function \cite{chen2016xgboost}.

      The hyper-parameters applicable are the regularization parameter ($\lambda$) and the gradient descent step-size / learning rate ($\alpha$).
  \end{itemize}
  
  The implementation library utilized for the Simple Linear Regression and Support Vector Regression techniques is Scikit-Learn \cite{pedregosa2011scikit}, whereas the XGB Python library was used for the XGBoost regression implementation.


\section{Training Dataset Augmentation} 
\label{sec:training_datasets_augmentation}

  A few different datasets were used to train the models on, in an attempt to identify the best representative training set.
  The dataset augmentation strategies used are enumerated below.

  \begin{itemize}
    \item
      \textbf{Article Content Expansion}

      To increase the number of features to train on, it was decided to retrieve the full text content of the articles corresponding to the article headlines. This was achieved by creating an application to search for the article headlines that were part of the training set using an online search engine, and to retrieve the full-text of the article by scraping the content from the source websites.

      This application is implemented in Java and is open-source\footnote{https://github.com/v1n337/news-article-extractor}. The implementation can be extended to augment any set of headlines with the corresponding article content.

      The assumption made here is that the sentiment expressed in the article headline sufficiently proxies the sentiment in the actual article content.

    \item 
      \textbf{Amazon Product Reviews} 

      This corpus is a set of Amazon product reviews\footnote{http://jmcauley.ucsd.edu/data/amazon/}, each consisting of the review text and a star rating on the scale of 1-5. To normalize the dataset, the rating scores 1 \& 2 are assumed to be associated with negative reviews, 3 with neutral and 4 \& 5 with positive reviews. This score range was then mapped to a -1 to 1 scale to match the sentiment scores of the training data. In total, $100,000$ documents from this dataset were used to augment the existing training dataset.
    \item 
      \textbf{Financial Phrasebank} 

      This dataset is specific to the financial domain and is manually annotated \cite{malo2014good}. It is comprised of a set of financial snippets from stock market related news that have been annotated with the classes positive, negative and neutral. 

      To normalize the labels, neutral was assigned a sentiment score of 0 and experiments were run for $positive \in (1, 0.5)$ and $negative \in (-1, -0.5)$. 
  \end{itemize}

None of the above strategies proved to be a good augmentation of the existing data, since their addition to the training datasets did not show any improvements in the overall cross-validated accuracy score.


\begin{table*}[t] 
  \centering
  \begin{tabular}{ | c | c |  c | } \hline
    \textbf{Vectorization Method} & \textbf{Learning Model} & \textbf{Cosine Similarity}\\ \hline 
    Unigrams \& Bigrams & Simple Linear Regression & 0.644 \\ \hline
    Unigrams \& Bigrams & XGBoost Regression & 0.547 \\ \hline
  \end{tabular}
  \caption{SemEval Task 5 Submissions}
  \label{tab:semeval_submission}
\end{table*}

\section{System Implementation} 
\label{sec:system_implementation}
  The entire system was coded in Python with the use of the Scikit-Learn \cite{pedregosa2011scikit}, XGB and Gensim libraries. This includes a framework for automated testing of accuracy scores to arrive at the best hyper-parameters to be used for unigram \& bigram word count combinations, as well as Doc2Vec hyper-parameters.

  The system implementation includes all the plugins pertaining to the different document vectorization techniques and statistical learning techniques discussed in sections \ref{sec:document_vectorization} and \ref{sec:regression_models} respectively.

  The code is open source\footnote{https://github.com/v1n337/semeval2017-task5} and is available to replicate the results published in this paper along with the instructions to operate the system. 

\section{Experimental Results} 
\label{sec:experimental_results}

    For arriving at the baseline scores, an exhaustive set of tests were conducted using each of the document vectorization techniques in combination with the regression techniques described in the previous sections. 

    Using the automated test-suite included as part of the system, it was concluded that the Doc2Vec model performed best when the number of dimensions (features) of text is around 832 and the learning algorithm completes 40 passes before settling on a vector representation. It was also concluded, that a combinations of unigrams \& bigrams had the best baseline accuracy scores for the training datasets.

    The measure of accuracy used was the $R^2$ score, also called the co-efficient of determination. The $R^2$ score can be computed using the below formula:
    $$R^2 = 1 - \frac{\sum_{i=1}^N (y_i - f_i)^2}{\sum_{i=1}^N (y_i - \bar{y})^2} $$
    where $y$ is the gold set score vector and $f$ is the predicted score vector, and $N$ is the number of test samples.

    The experimental results for the Training and the Trial datasets are shown in Table~\ref{tab:semeval_baselines}. The best baselines scores seem to favor the simplest vectorization model, i.e. unigrams \& bigrams.


\section{System Evaluation} 
\label{sec:system_evaluation}
  For the two submissions permitted by SemEval, the methods used for the submissions made are described in Table \ref{tab:semeval_submission}.

  The evaluation was done using the task evaluation metric, the $cosine\_score$ \cite{cortis-EtAl:2017:SemEval}.
  $$cosine\_score = cosine\_weight * cosine(G, P)$$
  where
  $$cosine(G, P) = \frac{\sum_{i=1}^N G_i * P_i}{\sqrt{\sum_{i=1}^N {G_i}^2} * \sqrt{\sum_{i=1}^N {P_i}^2}} $$
  and
  $$cosine\_weight = \frac{|P|}{|G|} $$
  and $G$, $P$ are the gold set scores and the predicted scores respectively, for $N$ test samples.
  
  The simplest model implemented, using Unigrams \& Bigrams, combined with Simple Linear Regression, was what yielded the best performance by the system, with a cosine similarity score of $0.644$.

\section{Conclusions and Future Work} 
\label{sec:conclusions_and_future_work}

  This paper has described the UW-FinSent system developed by the UWaterloo team for Task 5, Subtask 2 during SemEval 2017. 

  The experimental results indicate that the usage of simpler techniques like N-gram text vectorization and linear regression to predict the continuous-valued scores achieve better results than bag-of-words or deep learning feature extraction techniques.

  A recurring topic that needed to be addressed during the progress on this task was the fact there there were no reliable datasets that could accurately augment the training set. In the future, we plan to develop automatic methods for generating high quality, sentiment-annotated training datasets for the financial domain.


\section*{Acknowledgements} 
\label{sec:acknowledgements}

  The authors are grateful to the organizers for their support for this task. The authors would also like to thank \cite{malo2014good} for sharing the Financial Phrasebank Dataset for the purposes of our evaluation.


\bibliography{acl2017}
\bibliographystyle{acl-natbib}

\end{document}